# Reconsideration on evaluation of machine learning models in continuous monitoring using wearables

Cheng Ding[1,2], Zhicheng Guo[3], Cynthia Rudin[3,4], Ran Xiao[1], Fadi B Nahab[5], Xiao Hu[1,2,6]

[1]Nell Hodgson Woodruff School of Nursing, Emory University, Atlanta, GA, USA
[2]Wallace H. Coulter Department of Biomedical Engineering, Georgia Institute of Technology, Atlanta, GA, USA
[3]Department of Electrical and Computer Engineering, Duke University, Durham, NC, USA
[4]Department of Computer Science, Duke University, Durham, NC, USA
[5]Department of Neurology, Emory University School of Medicine, Atlanta, GA, USA
[6]Department of Biomedical Informatics, Emory University School of Medicine, Atlanta, GA, USA

**Abstract**

This paper explores the challenges in evaluating machine learning (ML) models for continuous health monitoring using wearable devices beyond conventional metrics. We state the complexities posed by real-world variability, disease dynamics, user-specific characteristics, and the prevalence of false notifications, necessitating novel evaluation strategies. Drawing insights from large-scale heart studies, the paper offers a comprehensive guideline for robust ML model evaluation on continuous health monitoring.

## 1. Introduction

The widespread adoption of electronic wearable devices with built-in biosensors has enabled their deployment to millions of users for various applications, including health condition monitoring, such as atrial fibrillation (AF) detection, blood pressure estimation, viral infections and blood oxygen saturation measurement [1-4]. Especially with the utilization of photoplethysmography (PPG) signal, these devices have demonstrated significant potential in providing real-time insights into an individual's health status. PPG, due to its non-invasive nature and ease of integration into wearable technology, has become a cornerstone in modern health monitoring systems [5]. Analyzing wearable device signals often involves ML models of different complexities [6, 7]. In the model development phase, typically, continuous signals are cut into discrete segments, and the model's performance is evaluated at the segment level using conventional metrics such as accuracy, sensitivity, specificity, and F1 score [8]. However, relying solely on these conventional metrics at the segment level does not provide a holistic assessment and hurts both consumers by making it impossible to select optimal solution for their needs and innovators by failing to guide their effort towards true progresses. The complex nature of continuous health monitoring using wearable devices introduces unique challenges beyond conventional evaluation approaches' capabilities, as illustrated in Figure 1. Recognizing these challenges is imperative for imbuing continuous health monitoring applications with accurate and reliable ML models to ensure a successful translation of these models into everyday use by millions of people and fulfill the potential of this technology at scale. In the subsequent sections, we outline the challenges in evaluating ML models for continuous health monitoring using wearables, thoroughly review existing evaluation methods and metrics, and propose a standardized evaluation guideline.

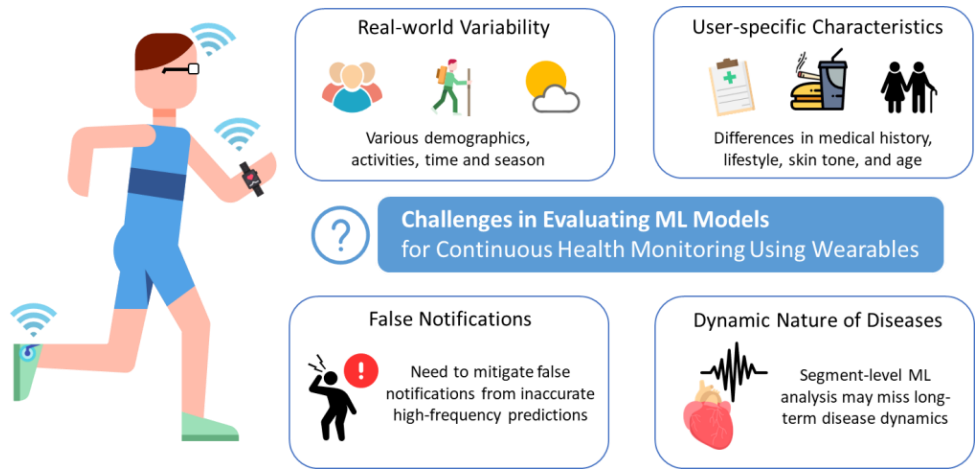

**Figure 1.** Challenges in Evaluating ML Models for Continuous Health Monitoring Using Wearables

## 2. Challenges in Evaluating ML Models for Continuous Health Monitoring Using Wearables

2.1 Real-World Variability and Environmental Factors

The quality of physiological signal can be impacted by many factors, including user activities, time of the day, ambient lighting conditions, electromagnetic interference, room temperature, humidity, and other user behaviors (e.g., compliance, comfort fit, etc.) [9]. These factors can lead to morphological changes in the collected signal, even with additional introduction of artifacts, affecting the ML model's performance and introducing uncertainty in predictions. For example, Fig. 2a shows the morphology of a young healthy person recovering from an intense cardio workout. Within a span of 20 minutes, the morphology has changed substantially. Fig. 2b shows a scenario where the morphology changes as one falls asleep over a span of around 30 minutes [10].

The challenge lies in creating evaluation guidelines that simulate and account for these real-world variations, ensuring that ML models can maintain their accuracy and robustness across diverse real-world scenarios. Importantly, achieving high accuracy in wearable data analysis from clean signals is not as impressive as achieving a high accuracy from signals corrupted by various real-world scenarios.

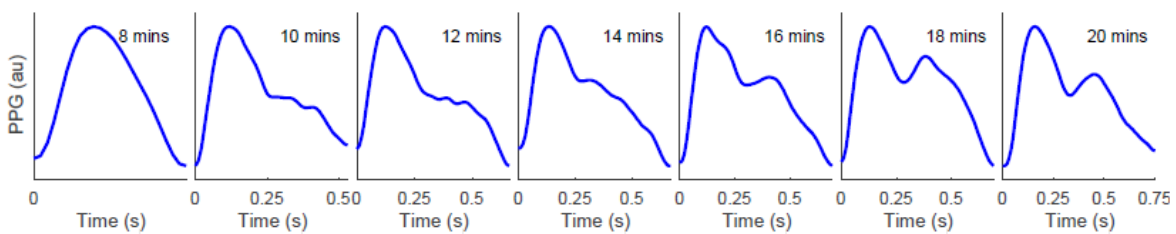

**Fig. 2a** Change in PPG morphology as one recovers from an intense cardio workout

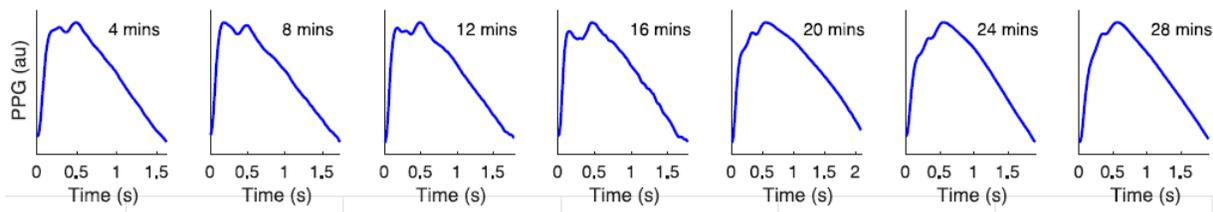

**Fig. 2b** Change in PPG morphology as one falls asleep. The figures are reproduced from [10] with permission from the authors under the Creative Commons license.

2.2 Dynamic Nature of Diseases

Segment-level analysis, such as breaking continuous data into 30-second intervals, offers a granular view of physiological changes. However, it may not effectively capture the prolonged and complex dynamics of certain diseases. For instance, some conditions, like early-stage AF, have intermittent manifestations that can easily go undetected with such short-term snapshots. Additionally, in diseases like AF, parameters like the AF burden – indicating the proportion of time the heart remains in AF – are pivotal in assessing associated risks [11, 12]. These crucial nuances might be overlooked when relying solely on segment-level metrics. Hence, while segment-level evaluation can provide valuable insights, it's essential to recognize its potential shortcomings in representing the comprehensive nature of certain diseases.

2.3 User-Specific Characteristics

User characteristics (e.g., demographics, medical history, etc.) add variability to the data. Often results are not reported over different subgroups of users but on the overall performance of the whole cohort. These users, especially patients, present a spectrum of physiological traits influenced by age, gender, skin tone, hair density, and underlying medical conditions, as shown in Figure 3. For instance, age stands out as a prominent risk determinant for AF, with its prevalence surging notably after age 65 [13]. Gender-wise, AF exhibits a higher incidence in men than women [14]. Notably, studies underscore the diminished accuracy of pulse oximeter-derived blood saturation (SPO2) measurements among users with darker skin tones as opposed to those with lighter skin tones [15, 16]. This is because the PPG amplitude is typically reduced due to higher absorbance by darker skin with higher melanin. In the context of ethnic variability, research reveals disparities in AF incidence among different ethnic groups; for instance, the MESA (Multi-Ethnic Study of Atherosclerosis) study reported lower AF incidence in Hispanics, Asians, and Blacks above 65 years when compared to non-Hispanic Whites. In addition, A larger study, encompassing over 600,000 patients from the Veteran Affairs, reported almost twofold higher age-adjusted AF prevalence in Whites as opposed to other ethnicities [17]. These variations highlight the importance of considering user characteristics in evaluating ML models to ensure their effectiveness and reliability across diverse populations.

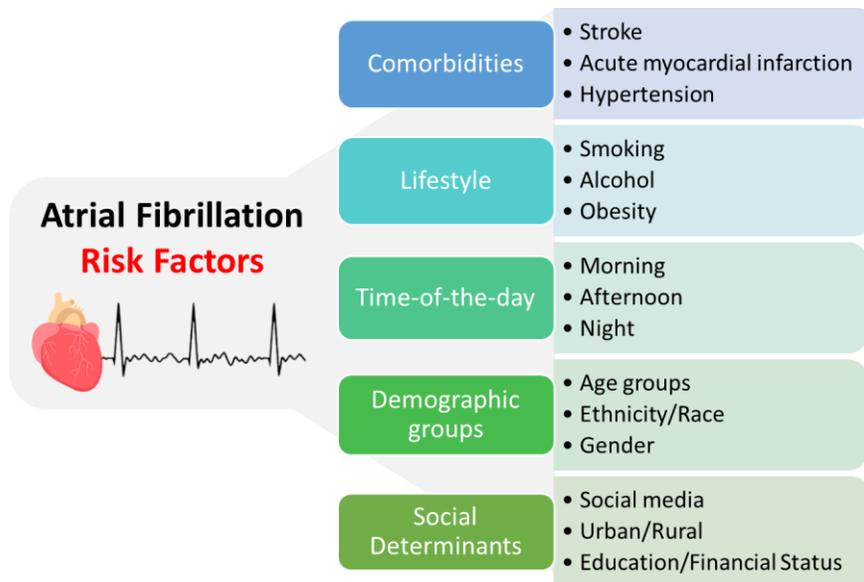

**Figure 3.** Different conditions and user characteristics that can influence ML model performance

2.4 Management of False Notifications

False notifications can cause anxiety and alarm fatigue. The continuous nature of the data generated from wearable devices results in a high frequency of predictions, often producing hundreds of predictions per hour when using relatively short segment lengths (e.g., 30 seconds). Even a model with over 90% accuracy can generate massive false predictions. Navigating this challenge requires aggregating segment level detections and, hence, an evaluation approach that extends beyond using segment-level metrics.

## 3. Current Status of Evaluation of Continuous Health Monitoring in Consumer-Level Products: Insights from Large-Scale Heart Studies

Besides ML model development in academic communities, large-scale heart studies have been conducted by big technology companies for their consumer-level wearable products, including Apple, Fitbit, and HUAWEI [7, 18, 19]. Although their ML models analyze segments, notifications about a possible medical event are generated only when specific criteria based aggregations of segment level results are met. For example, in the Apple Heart study [18], a 1-minute PPG tachogram was calculated every two hours. Five consecutive irregular tachograms within 48 hours generated a notification to users. In the Fitbit Heart study [19], 5-minute tachograms were collected every 2.5 minutes (i.e., overlap of 50%). An irregular heart rhythm notification was triggered once 11 successive irregular tachograms were identified. In the HUAWEI Heart study [7], evaluation was done every ten minutes. A notification was generated for users once greater than ten measurements were identified as AF in 24 hours.

All these heart studies have drawbacks. First, tradeoffs are essential between high precision and high sensitivity for screening diseases at a population level. For example, in the Apple Heart study, in the group of users who received irregular heart rhythm notifications (n=929), 404 reported a new AF diagnosis. At the same time, 3070 out of 293,015 users who did not receive irregular heart rhythm notifications also had new AF diagnoses according to the end-of-study survey. Thus, the sensitivity of AF detection was only 11.6%. Second, studies did not report if the AF condition was detected promptly to start an intervention. Third, ML models were evaluated on data where subjects were stationary, where evaluation is much easier than in realistic settings. For example, the Apple Heart study stated: 'Tachograms are collected and analyzed only if the user is still enough to obtain a reading.' In Fitbit's Heart study, the PPG data was analyzed only during stationary periods, where accelerometer data determined stationarity. HUAWEI's Heart study states that 'active measurement' was only performed while the user was at rest. Thus, in large-scale consumer wearable health studies, the evaluation strategies for disease detection should be further refined. Additionally, all the aggregation strategies that were used in the studies (e.g., 10 irregular measurements in 24 hours) are heuristics and do not appear to be carefully optimized for optimal detection of AF.

Importantly, we note that none of the data from these studies are available for the purpose of reproducibility or secondary analysis [20].

## 4. The Need for a Structured guideline when Evaluating ML Models for Continuous Health Monitoring Using Wearables

This perspective paper advocates the importance of establishing a ML model evaluation guideline tailored to address the inherent challenges with the continuous nature of health monitoring with wearable devices. Many different entities can benefit from our guideline. Regulatory bodies such as the U.S. Food and Drug Administration (FDA) could benefit by adapting and incorporating our guideline when assessing submissions of continuous health monitoring solutions. Device vendors can also leverage our proposed evaluation guideline to improve the efficacy and safety of their solutions. Academic researchers can use the establishment of consensus evaluation metrics within their studies to foster a standardized approach to assessing ML models' effectiveness, promoting transparency and comparability across different studies. Moreover, if users can assess wearable devices using detailed performance reports, it would allow them to make informed decisions and select the most suitable product for their specific health needs and preferences.

## 5. A Guideline for Reporting ML Models on Wearable Devices for Health Monitoring

Ensuring the reliability and efficacy of ML models for health monitoring via wearable devices necessitates a meticulous approach to development, evaluation, and reporting. To that end, we propose a guideline for reporting the performance of ML models in wearable devices for health monitoring to facilitate robust, reliable, and ethically deployed ML models. The guideline consists of four main areas, as illustrated in Figure 4.

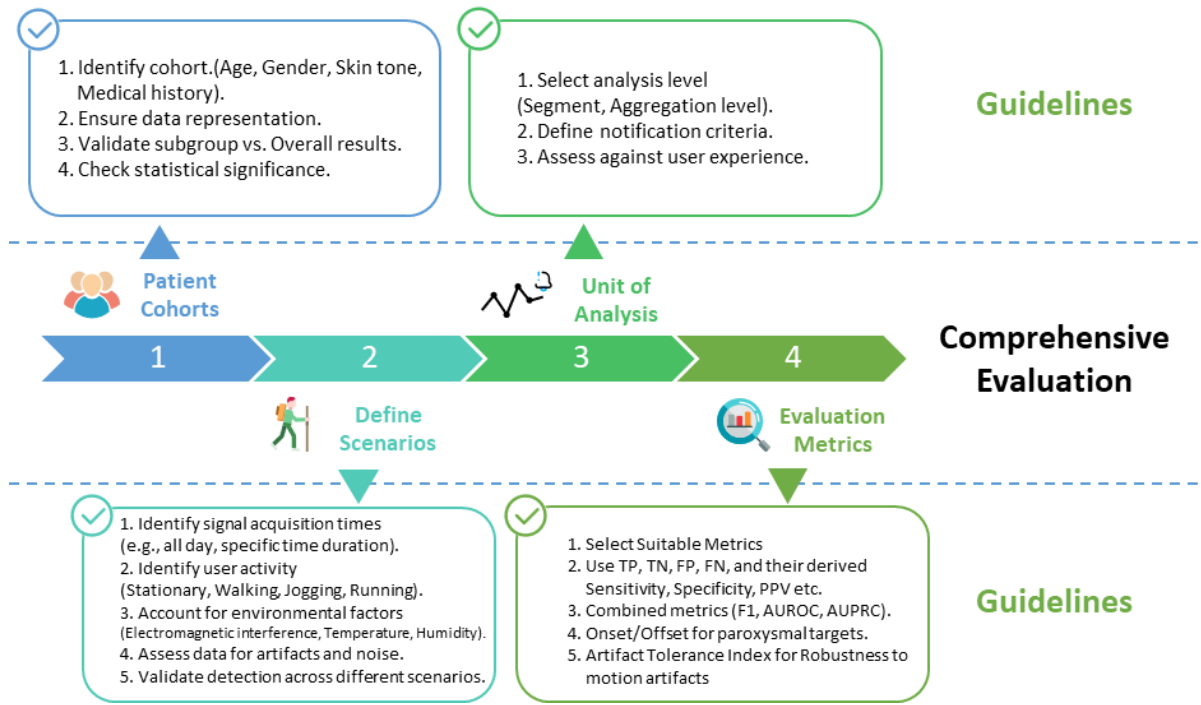

**Figure 4**. A Guideline for Reporting ML Models on Wearable Devices for Health Monitoring

5.1 Cohort Selection

The evaluation starts with providing a descriptive analysis of the study cohort. For models targeting the general population with consumer-grade wearables, we suggest the study cohort be composed of a broad and diverse participant distribution, with respect to factors such as age, gender, and race. On the other hand, for models targeting a specific patient population, the study cohort becomes more focused than that of the general population but still needs to consider the demographic and biological factors to ensure an ethical and equitable model.

5.2 Define the Target Scenarios

As the second step, defining target scenarios involves identifying and outlining specific use cases under which the ML models are expected to function effectively. These scenarios might encompass various states of physical activity, environmental conditions, and physiological states. For instance, in the context of wearable devices for health monitoring, the model must demonstrate reliability across scenarios like active exercise, sleep, and sedentary behavior, each presenting unique challenges in terms of data quality and contextual relevance. Specific attention must be given to formulating a rationale for each scenario, ensuring that each is pertinent to the potential user base and aligned with real-world deployment contexts.

5.3 Evaluation Approaches and Notification Strategy

A notification strategy must be devised that balances the user experience and clinical urgency. As shown in Figure 5, the **segment-level analysis** imposes a high demand for model accuracy to reduce false alerts. While aggregation-level analysis reduces temporal resolution, it permits the evaluation of ML models over extended time units, such as on an hourly or daily basis. This approach involves aggregating segment-level predictions to generate predictions for larger time intervals. It also serves as a filtering mechanism and helps reduce sporadic alerts which can be false positives.

To effectively assess the performance of various aggregation strategies, the precision-recall curve is a pivotal metric. By plotting precision against recall, this curve offers a visual representation of the trade-offs between false alerts and missed events inherent to each strategy. The area under the curve (AUPRC) serves as an aggregate measure of this performance. By comparing the AUPRC values for different aggregation methods, one can discern which approach best navigates these trade-offs. Such a metric becomes especially vital in contexts with imbalanced datasets, ensuring the selected aggregation strategy offers a balanced and reliable output.

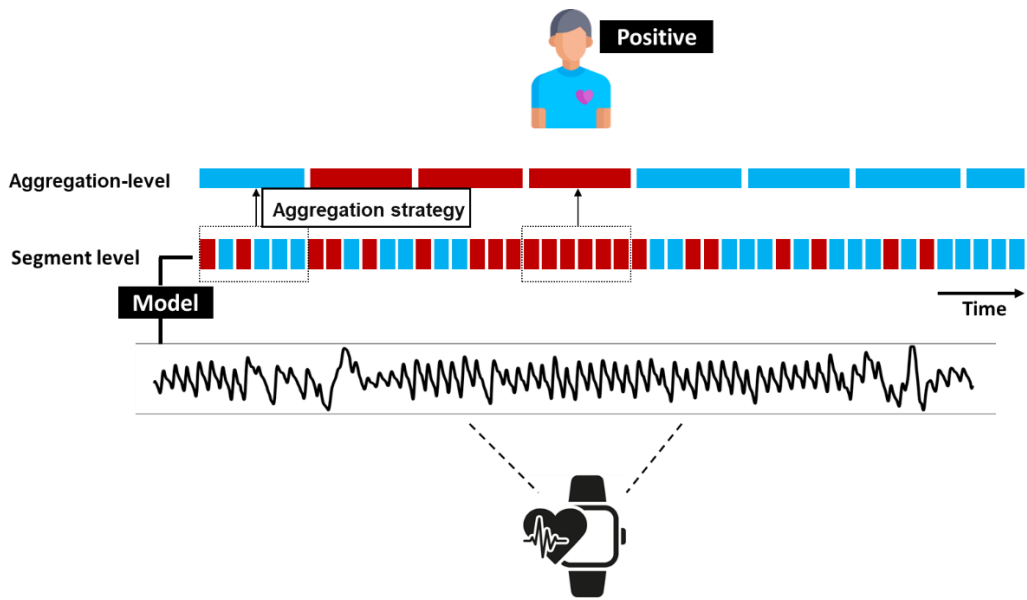

**Figure 5.** Performance evaluation with different units

5.4 Metrics Selection

The performance evaluation expands beyond conventional metrics like AUROC, accuracy, specificity, and sensitivity and incorporate factors salient to health outcomes, such as time-to-detection for critical events or robustness to data artifacts (e.g., sensor noise during physical activity). We are introducing several metrics which are suitable to be considered in a continuous manner.

***Onset* and *offset* estimation.** We propose that ML models be evaluated on accurate detection of the *onset* and *offset* of a condition. *Onset* and *offset* are more semantically meaningful for designing user-centric notifications. For example, a user may be more interested in knowing the *onset* when the goal is early detection, while another user may be more interested in knowing the *offset* post-treatment. *Onset* and *offset* of the condition can also facilitate investigating potential causal relationships between a health condition and external triggers such as consumption of alcohol or coffee for arrhythmias [21, 22]. Accurate estimation of *onset* and *offset* are important for deriving additional metrics, such as **condition burden** (or its counterpart, percentage of condition-free time) that quantify the **frequency, duration, and severity of disease** episodes. This estimation holds clinical significance as it aids in understanding the overall impact on an individual's health, potentially influencing treatment decisions and intervention strategies. Accuracy in *onset* and *offset* estimates is different from conventional metrics that predominantly focus on isolated events.

**Artifact Tolerance Index.** To gauge the robustness of an ML model in the presence of artifacts, we propose a novel metric termed the "Artifact Tolerance Index" (ATI). To calculate it, we first identify the percentage of artifacts within each segment using SQIs [23, 24]. We then calculate ATI, which quantifies the ML models' ability to remain accurate and reliable across the full range of artifact levels within the collected data. This can yield an artifact tolerance curve (ATC) as shown in Fig. 6, where the x-axis denotes the percentage of good quality segments (i.e., 1 – artifact coverage) and the y-axis denotes a desired performance metric. The curve demonstrates how the model's performance degrades with increased artifact percentages. ATI can be estimated in different ways. One way is to calculate it as the slope of the performance deterioration vs. artifact percentage curve. A larger slope indicates less resilience of the model to signal artifacts, and vice versa. Another potential way to estimate ATI is to fix a minimum performance requirement and then report the highest artifact coverage (or smallest SQI) for which that performance can be maintained. Lastly, one can calculate the area under the artifact tolerance curve (AUATC) as an overall indicator to compare different models.

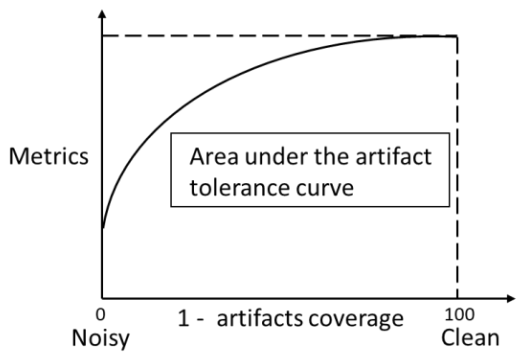

**Figure 6.** Performance change as a function of artifact coverage to estimate ATI as the area under the artifact tolerance curve.

## 6. Conclusion

This perspective highlights the critical need to refine the evaluation methodologies for ML models in the realm of continuous health monitoring through wearable technologies. Beyond conventional metrics, we emphasize the significance of accurate onset and offset estimation for the target clinical endpoint. We also shed light on the challenges posed by real-world variability, disease dynamics, user-specific attributes, and the presence of false notifications, underscoring the need for innovative approaches to performance evaluation. Moreover, this perspective highlights the ethical mandates of bias mitigation, transparency, and post-market surveillance to ensure wearable monitoring technologies' ethical and equitable deployment. This work advocates for evaluating wearable monitoring models through more comprehensive evaluations, thus promoting enhanced healthcare support and user confidence in the digital health landscape.

## Acknowledgment

This work was partially supported by NIH grant award R01HL166233. We also greatly appreciate the editorial contributions from Professor Roy L. Simpson.